\documentclass[runningheads,a4paper]{llncs}

\usepackage{amssymb}
\usepackage{graphicx}
\usepackage{multirow}
\usepackage{cite}
\usepackage{subfigure}
\usepackage{booktabs}
\usepackage{url}

\urldef{\mailsa}\path|{yhuangch, achung}@cse.ust.hk|    
\newcommand{\keywords}[1]{\par\addvspace\baselineskip
\noindent\keywordname\enspace\ignorespaces#1}

\begin{document}

\mainmatter  % start of an individual contribution

\title{Improving High Resolution Histology Image Classification with Deep Spatial Fusion Network}

\titlerunning{Deep Spatial Fusion Network}

%\author{Yongxiang HUANG\and Albert Chi-Shing CHUNG}
\author{Yongxiang HUANG%
%\thanks{}%
\and Albert Chi-Shing CHUNG}
%

\institute{Lo Kwee-Seong Medical Image Analysis Laboratory, \\
Department of Computer Science and Engineering,\\
The Hong Kong University of Science and Technology, Hong Kong\\
\mailsa\\}

\toctitle{}
\maketitle

\begin{abstract}
Histology imaging is an essential diagnosis method to finalize the grade and stage of cancer of different tissues, especially for breast cancer diagnosis. Specialists often disagree on the final diagnosis on biopsy tissue due to the complex morphological variety. Although convolutional neural networks (CNN) have advantages in extracting discriminative features in image classification, directly training a CNN on high resolution histology  images is computationally infeasible currently. Besides, inconsistent discriminative features often distribute over the whole histology image, which incurs challenges in patch-based CNN classification method. In this paper, we propose a novel architecture for automatic classification of high resolution histology images.  First, an adapted residual network is employed to explore hierarchical features without attenuation. Second, we develop a robust deep fusion network to utilize the spatial relationship between patches and learn to correct the prediction bias generated from inconsistent discriminative feature distribution. The proposed method is evaluated using 10-fold cross-validation on 400 high resolution breast histology images with balanced labels and reports 95\% accuracy on 4-class classification and 98.5\% accuracy, 99.6\% AUC on 2-class classification (carcinoma and non-carcinoma), which substantially outperforms previous methods and close to pathologist performance. % We provide our implementation on github[] and are sure that the proposed architecture can be easily applied to other high resolution medical images. 

\end{abstract}

\section{Introduction}

Histology imaging on tissue slice is a critical method for pathology analysis, indicating further targeted therapies. Pathologists perform histological analysis and morphological assessment on microscopic structure and tissue organization to diagnose and grade the cancer type \cite{4rosen2001rosen}. On histology images, discriminative features of a certain cancer type can be observed at nuclei-level, ductal-level, cellular-level and overall tissue organization \cite{5gurcan2009histopathological}. The diagnosis of biopsy tissue is tedious and non-trivial. In-observer disagreement often exists between pathologists due to the complex diversity and distribution of discriminative features \cite{2elmore2015diagnostic}. %Here we should mention the in-observer disagreement rate between pathologists. Take breast cancer for example
Therefore, developing an accurate computer-aided diagnosis (CAD) system to automatically classify histology images can greatly improve diagnosis efficiency and provides valuable diagnosis reference to pathologists with dissensions [\cite{1yamada1993computer}].

In recent years, deep convolutional networks have achieved state-of-the-art performance on a large number of visual classification tasks  {\cite{19he2015delving}}. The success of deep CNN relies on the large available training set that is well labeled and is limited to the size of input image considering the high computational cost. However, for classification problems in biomedical images, the input is often high resolution images, such as breast cancer histology images. 

The challenges of developing a CNN for high resolution histology images classification include: (1) the distribution of discriminative features over a histology image is complex and one patch on a histology image does not necessarily contain discriminative features consistent with the image-wise label; (2) In most cases, only the image-wise ground truth label is given due to the high cost of annotation on high resolution images, which complicates the problem; (3) Dramatic down-sampling leads to the loss of discriminative details at nuclei-level and ductal-level, thus training a CNN on whole histology images is usually inappropriate. \cite{6araujo2017classification} proposed to divide a high resolution breast histology image into patches and train a VGG-like patch-wise CNN, from which the image-wise label can be inferred by voting on patch-wise predictions. \cite{21rakhlin2018deep} utilized deep CNNs for feature extraction and gradient boosted trees for classification, which achieved better performance. \cite{6araujo2017classification} developed a patch-based CNN and utilized a linear regression fusion model to predict image-wise labels. 

Compared to previous methods, in this paper, we propose to use a deep spatial fusion network to model the complex distribution of discriminative features over patches. We also investigate a more effective method to extract hierarchical discriminative features on patches. The proposed method outperforms previous state-of-the-art solutions and  it is the first work that utilizes the spatial relationship between patches to improve image-wise prediction for breast cancer histology image classification. 

\section{Methods}

\subsection{Architecture}

\begin{figure}
\centering
\includegraphics[width = \linewidth]{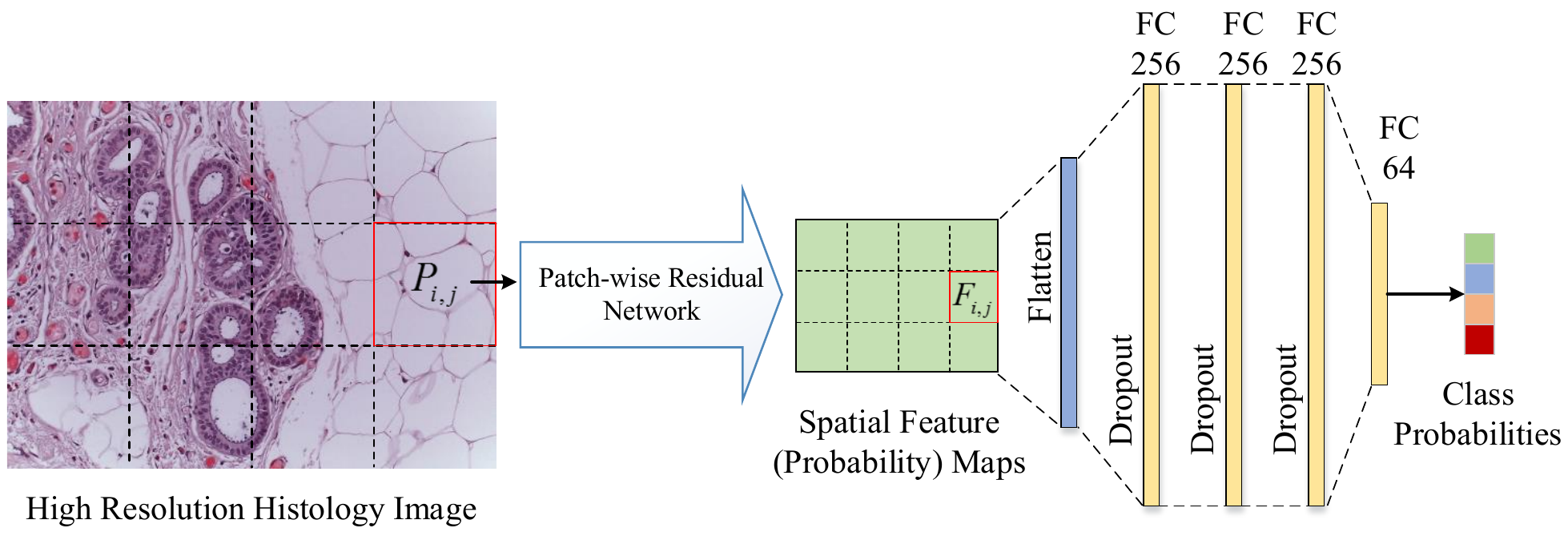} 
\caption{A schematic view of the proposed Spatially Fused Residual Network. The high resolution histology image is sampled into patches by a non-overlap sliding widow. $F_{ij}$ represents deep CNN features for a patch $P_{ij}$, where $i,j$ are the patch row and column index, respectively.}
\label{fig:arch}
\end{figure}

The architecture of the proposed method is depicted in Fig.{~\ref{fig:arch}}.  The input to the network is high resolution histology images. As discussed in Section 1, only image-wise ground truth labels are available and discriminative features may distribute sparsely on the whole image, which indicates that all patches are not necessarily consistent with image-wise label. We propose a spatial fusion network to model this fact and produce a robust image-wise prediction. 

The architecture is composed of two principal components: (1) an adapted deep residual network trained to discover hierarchical discriminative features  and predict the probabilities of different cancer type for local image patches. Compared to VGG-like convolutional network \cite{7simonyan2014very}, the skip connection structure of residual network reduces the vanishing gradient phenomenon in backpropagation and thus have a better performance on extracting critical visual feature with a deeper convolutional network. (2) a deep spatial fusion network, which is designed to utilize the spatial relationship between patches with the input of spatial feature maps. For the sake of simplicity and generalization, patch-wise probability vector is adopted as the base unit of the spatial feature maps as shown in Fig.~\ref{fig:arch}. The fusion model learns to correct the bias of patch-wise predictions and yields robust image-wise prediction compared to typical fusion methods, which will be discussed in the paper.

\subsection{Patch-wise Residual Network }
\label{2.2}

%and receptive filed illustrated in Table 1

Residual neural networks (ResNets) \cite{8he2016deep} are adopted in our proposed architecture instead of plain feedforward deep convolutional neural network. Compared to plain CNN, residual networks mitigate the difficulty of training deep network using shortcut connections and residual learning \cite{8he2016deep}. Additionally, the identity shortcut connections enable flow of information across layers without attenuation caused by non-linear transformations. Therefore, hierarchical features from the low level to  higher level are combined to make the final prediction, which is very useful in histology image classification considering discriminative features are distributed in the image from the cellular level to tissue level. %For example, to identify breast histology images, pathologists identify the cancer type by observing features at different scales including both nuclei and overall tissue organization \cite{5gurcan2009histopathological}. 

\begin{figure}
\centering
\vspace{-0.4cm}
\includegraphics[width = \linewidth]{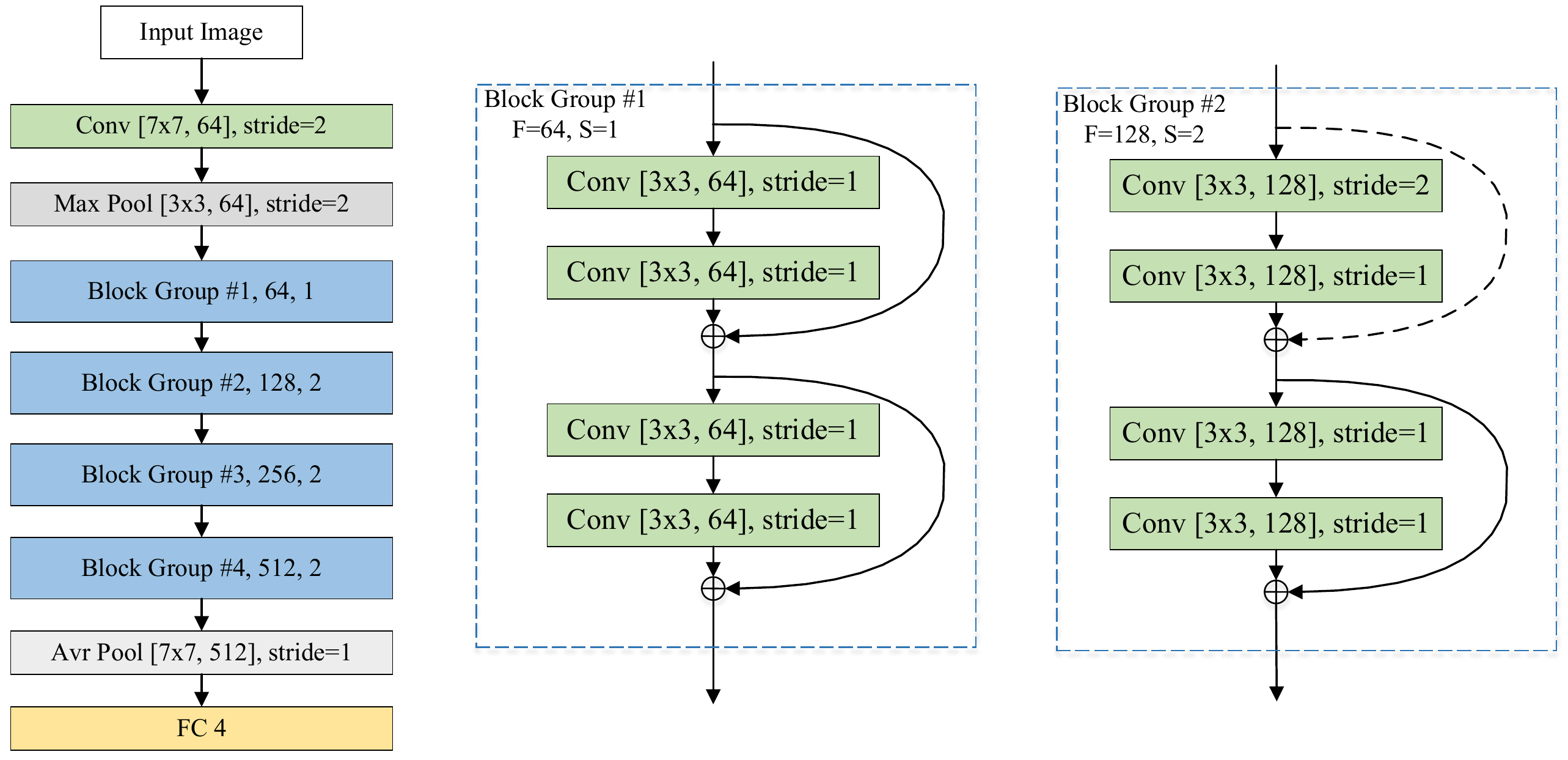} 
\vspace{-0.6cm}
\caption{The structure of the adapted residual network. Parameters in green boxes indicate "[kernel size, channels], stride". Each convolution layer follows by batch normalization and ReLU for regularization and non-linearity. Each block group consists of two building blocks \cite{8he2016deep}. The parameters in blue boxes indicate "\#index, the number of output feature map in each convolutional layer of the same group, stride of the first convolutional layer". In block group \#1, only identity shortcut connections are used because the input and the output are of the same dimensions. In block group \#2, both projection shortcut (dotted line) and identity shortcut are used. The projection shortcut matches the dimensions of the input and the output by $1\times1$ convolution. Block group \#3 and \#4 share structure with block group \#2.
	}
\label{fig:resnet}
\end{figure}

We have adjusted the original residual network developed for ILSVRC2015 classification task so that it works more appropriately on histology images classification. The modified residual network architecture is shown in Fig.~\ref{fig:resnet}. (1) The input layer is adjusted to receives normalized image patches of size $512 \times 512$, sampled from whole histology images. (2) The depth of the network is chosen to be 18 layers with 4 block units for fully exploring regional patterns in different scale. The receptive fields of the four block groups are of size 19x19 to 43x43, 51x51 to 99x99, 115x115 to 211x211, and 243x243 to 435x435 pixel respectively, which effectively respond to region patterns in nuclei, nuclei organization, structure and tissue organization \cite{6araujo2017classification}.

\subsection{Deep Spatial Fusion Network}

The aim of fusion model is to predict the image-wise label $\hat{y}$ among $K$ classes $\mathbf{C}=\{C_1,C_2,..,C_K\}$, given all patch-wise probability (feature) maps  $ \mathbf{F}$ output by the proposed residual network. The image-wise label prediction is defined by MAP estimate \cite{20greig1989exact} as follows,
\begin{equation}
  \hat{y} = \textrm{argmax}_{y \in{\mathbf{C}}}  P(y|\mathbf{F})  .
\end{equation}
 
Suppose the whole high resolution image is divided into $M \times N$ patches. We first organize all patch-wise probability maps in spatial order, such that:
\begin{equation}
  \mathbf{F} =
\left( \begin{array}{cccc}
F_{11} & F_{12} & \ldots & F_{1N} \\
F_{21} & F_{22} & \ldots & F_{2N}\\
\vdots & \vdots & \ddots & \vdots \\
F_{M1} & F_{M2} & \ddots & F_{MN}
\end{array} \right).
\end{equation}

 A deep neural network (DNN) is applied to utilize the spatial relationship between patches. As shown in Fig.~\ref{fig:arch} , the proposed fusion model consists of 4 fully-connected layers, each of which follows by ReLU activation function \cite{9nair2010rectified}. The deep multi-layer perceptron(MLP) learns to transform the spatial distribution of local probability maps to a global class probabilities vector during image-wise training. To increase the robustness of the model and avoid overfitting, we insert one dropout layer before each hidden layer. Notice that dropout layer is also inserted between the flatten probability maps and the first hidden layer. Dropping out half of the probability maps,  the models tend to yield an image-wise prediction with half information of patches through minimizing the cross-entropy loss in training.

\section{Experiments}

\subsection{Dataset and Preprocessing}
We validate the proposed method on two public histological breast cancer images dataset, the Bio-imaging Challenge 2015 Breast Histology dataset (BIC) \cite{url2}, and the BACH 2018 dataset \cite{url1}. Both datasets consist of Hematoxylin and Eosin (H\&E) stained microscope histology images on breast tissue biopsy. The images are annotated by two pathologists and classified into 4 classes: (1) normal tissue, (2) benign lesion, (3) in situ carcinoma, (4) invasive carcinoma, according to the predominant cancer type in each image. Among the 4 classes, in situ carcinoma and invasive carcinoma fall to  malignant carcinoma. As some works focus on analysis for malignant-benign classification, we also evaluate the performance of the proposed model on 2-class classification in our experiments. 

The BIC dataset consists of 286 high resolution images of size $2048\times1536$ pixels, split into 249 for training and 36 for testing. The BACH dataset consists of 400 images of the same size, split into 360 for training and 40 for testing. For both dataset, the 4 class labels are evenly distributed, hence it is fair to use accuracy as the evaluation metric. To avoid overfitting due to the small training dataset, we perform strong data augmentation as described in the next subsection. Before augmentation, to reduce the variance incurred by H\&E staining, the images are normalized using the method proposed in \cite{17macenko2009method}. %The colors of images are converted to optical density(OD) by logarithmic transformation and then singular value decomposition(SVD) is applied to OD tuples to find the 2D projections with high variance. The resulting color space transform is then applied to the original image.

\subsection{Network Training}

We first extract 512x512 pixel patches with overlapping from the high resolution images, as the input for the patch-based deep model. As patch-wise labels are not given in the training dataset, we initially assume the patch labels are consistent with the image-wise ground truth. It may incur bias during patch-based training and reduce the patch-wise classification accuracy. However, the bias will be alleviated during image-based training in the second stage under the supervised learning of image-wise labels. Due to the limited number of training samples, to prevent overfitting, we perform three kinds of image augmentation in each iteration:  (1) random rotation; (2) horizontal flipping; (3) random enhancement of contrast and brightness. Thus, we generate 201,600 patches from the BACH dataset and 140,000 patches from the BIC dataset respectively. The residual network is trained on 32-sized mini-batches to minimize the cross-entropy cost function using Adam Optimization \cite{22kingma2014adam} with learning rate $10^{-3}$ for 50 epoch. After training, the patch-wise network encodes a 512x512 patch to 10x10 feature maps and 4-class probabilities.%We compare two training strategies: transfer learning from ImageNet and training from scratch. For transfer learning, we incorporate the weights pre-trained on the ImageNet excepts for that of the last two layers as the initial weights for our residual network.  

To train the spatial fusion network, we perform similar data augmentation and generate 5,760 high resolution images from the BACH training dataset and 3,984 from the BIC training dataset. After augmentation, each high resolution image is divide into 12 non-overlapping patches of size 512x512 pixels. Each patch is fed into the residual network separately and output 512 feature maps of size 10x10 and a class probability vector of size 1x4. Probabilistic vectors of patches in the same image are then combined into a probabilistic map following their spatial order, which becomes the input for the spatial fusion network. The generated probability map can be seen as a high-level feature map that encodes all the patch-wise discriminative features and the image-wise spatial context-aware features. Supervised by the image-wise ground truth, the weights of the spatial fusion network are learned by using mini-batch gradient descent (batch size 32) with Adam optimization. During training, to minimize the cross-entropy loss, the spatial fusion model learns to encode the biased probabilistic map into a k-class vector approximating the image-wise ground truth (k=4). By utilizing the spatial context-aware feature hidden in the probabilistic map, the image-based classification accuracy can be effectively improved.

\subsection{Results}

We first evaluated the performance of the patch-based residual network and then focused on the effectiveness of the proposed spatial fusion network by conducting multiple comparison experiments on the two datasets. In the first experiment, we reimplemented the published state-of-the-art framework \cite{6araujo2017classification} on the BIC dataset as the baseline method, which is based on a patch-based plain CNN architecture followed by multiple vote-based fusion strategies (named Baseline). For a fair comparison, we only replaced the plain CNN architecture with the proposed patch-wise residual network (named ResNet + Vote) and evaluated the two methods using the same dataset setting. The results are shown in Table.{~\ref{table1}}. Residual + Spatial Network is our proposed method, which is evaluated on two dataset for a comprehensive comparison. All methods are evaluated with stratified ten-fold cross-validation on the same released dataset respectively.

\begin{table}[h]
    \centering
    %\vspace{-10pt}
    \caption{Quantitative comparisons on two public datasets. 
    % The baseline method is a published state-of-the art framework \cite{6araujo2017classification} on the BIC dataset, which is based on a plain CNN architecture followed by a voting strategy. ResNet + Vote uses the same two-stage framework but replaces the plain CNN architecture with our proposed residual network. ResNet + Spatial Fusion is our proposed method.
    }
    \begin{tabular}{l l c c c}
        \toprule
        Datasets	&   Methods 		& 4-class ACC 	&2-class ACC & STD\\
        \midrule    
        BIC		&	Baseline 		&0.778	& 0.833	& -\\
        BIC		&	Residual + Vote	&0.816	&0.850	& -\\
        BIC		&	Residual + Spatial Network	&\textbf{0.861}	&0.889 & -\\
        BACH	&	CNNs + GDT 		&0.872	&0.938 & 0.026\\
        BACH	&	Residual + Spatial Network 	&\textbf{0.950}	&0.985 & 0.022\\

        \bottomrule
    \end{tabular}
    %\vspace{-10pt}
    \label{table1}
\end{table}

\begin{figure}
\centering
\subfigure[Confusion matrix]{
\includegraphics[scale=0.35]{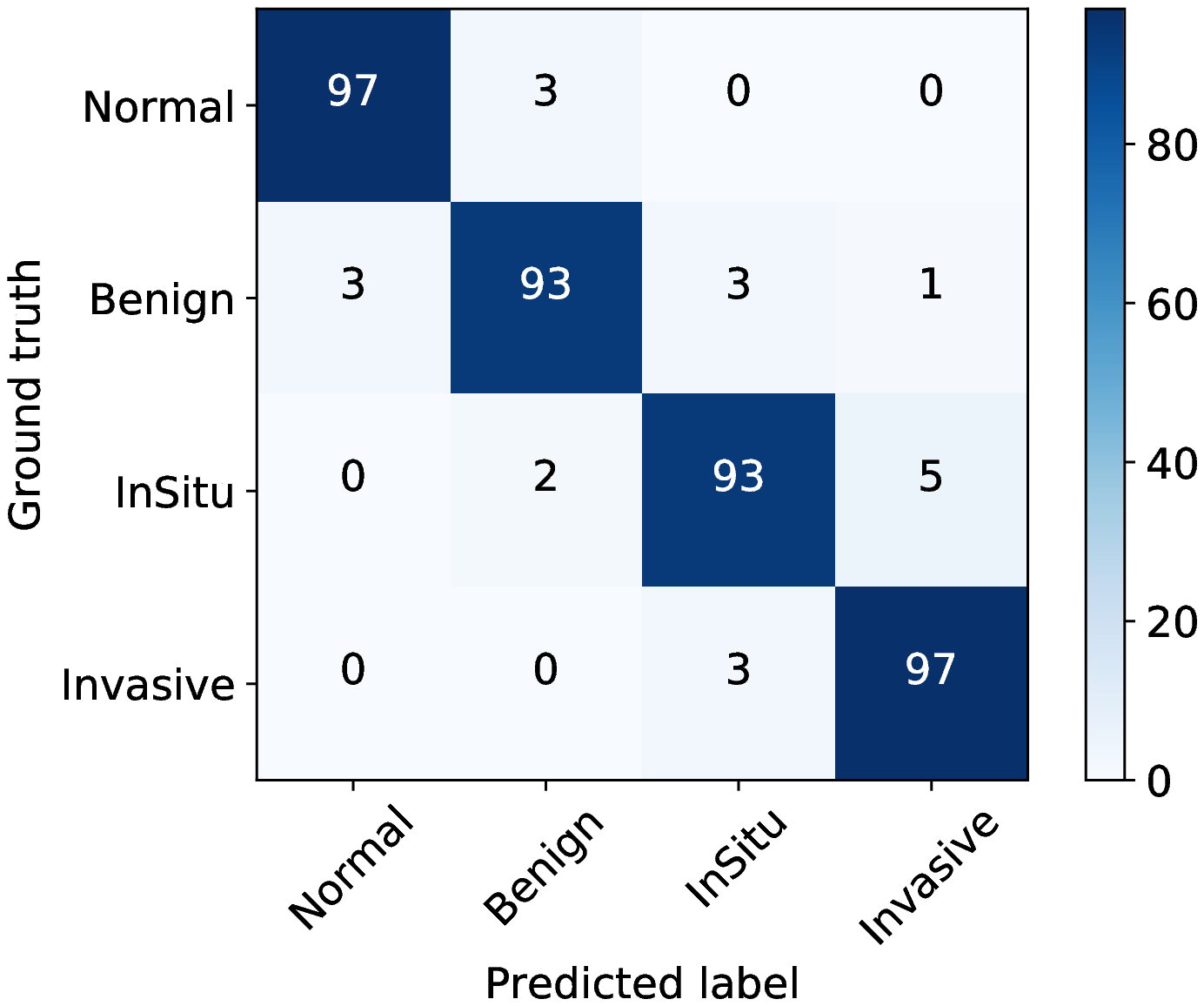} %width=\linewidth 
}
\subfigure[ROC]{
\includegraphics[scale=0.36]{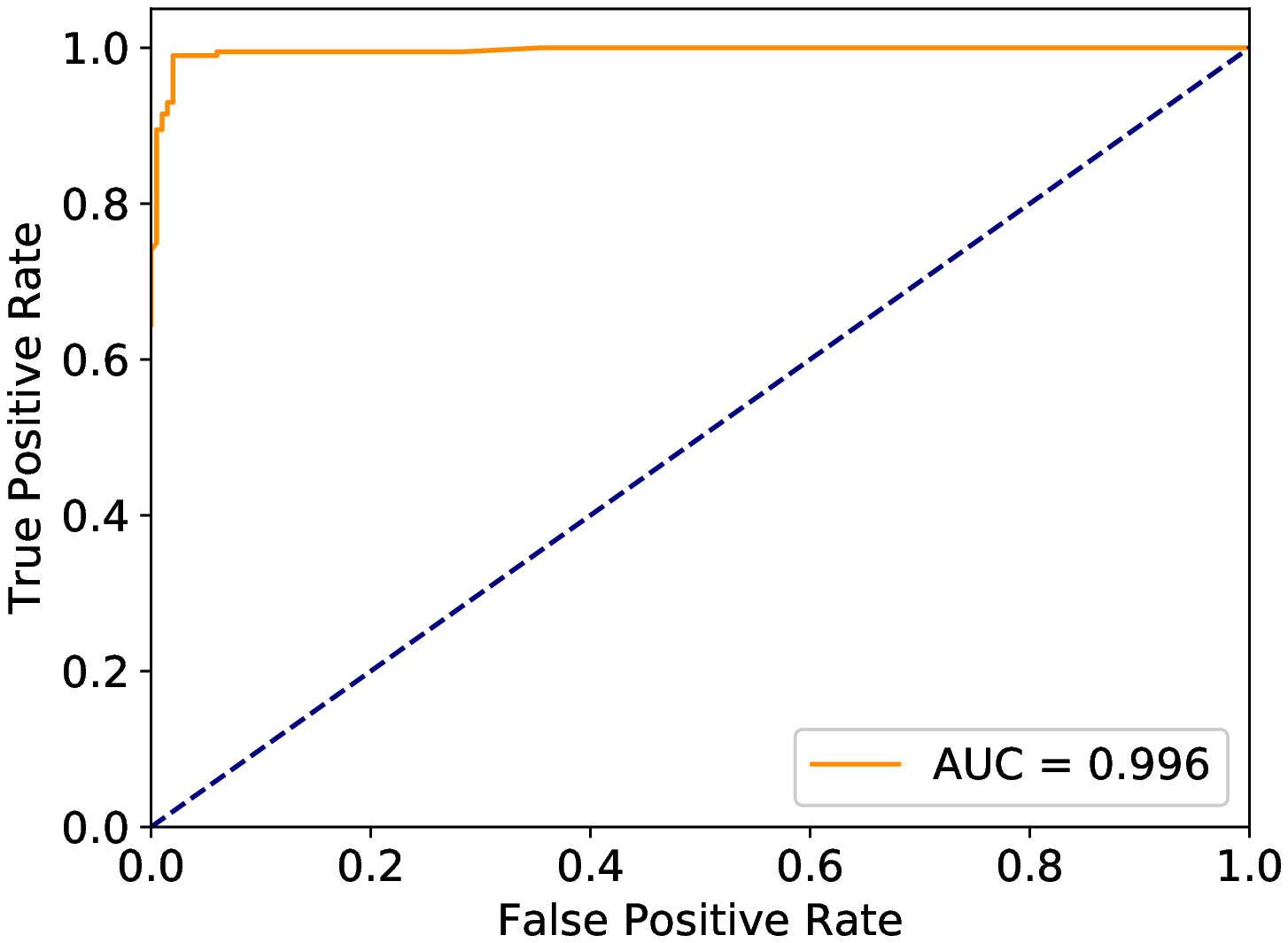} %width=\linewidth
}

\vspace{-8pt}
\caption{ (a) Confusion matrix without normalization, representing the 10-fold cross-validation result on 4-class classification of 400 high resolution histology images. (b) Performance of 2-class classification (non-carcinoma and carcinoma) in terms of AUC} 
\label{fig:cm4}
\end{figure}
\vspace{-10pt}

On the BIC dataset, the proposed method reports an accuracy of 86.1\% for 4-class classification, which outperforms the baseline method \cite{6araujo2017classification} by 8.31\% . The proposed patch-wise residual network brings an improvement of 3.8\% by replacing the plain CNN in the baseline method. The deep spatial fusion network further improves the ResNet + Vote method by 4.51\% further, which demonstrates utilizing the spatial context-aware feature map is more effective than using multiple voting strategies for patch-wise result fusion. On the BACH dataset, the proposed method reported 95.0\% accuracy on 4-class classification and 98.5\% accuracy, 99.6\% AUC on 2-class classification (carcinoma and non-carcinoma). As a comparison, CNNs + GDT \cite{21rakhlin2018deep} is a published state-of-the-art method on the BACH dataset, which adopted   several deep CNNs (ResNet50, InceptionV3, and VGG16) by model ensemble and used gradient boosted trees classifier to extract features at different scales. The proposed spatial fusion network outperforms \cite{21rakhlin2018deep} by 7.8\% on 4-class classification without using any ensemble technique. The classification performance in terms of confusion matrix and  receiver operating characteristic curve (ROC) are shown in Fig.{~\ref{fig:cm4}}, .  

%The proposed method reported 95.0\% accuracy on 4-class classification and 98.5\% accuracy, 99.6\% AUC on 2-class classification (carcinoma and non-carcinoma). \cite{21rakhlin2018deep} utilized strong data augmentation and adopted several deep neural network (ResNet50, InceptionV3, and VGG16) and gradient boosted trees classifier to extract features at different scales, which achieves a state-of-art performance on the BACH dataset. Our method significantly outperforms it with a 7.8\% improvement on 4-class classification accuracy without using ensemble. On the BIC dataset, the proposed method outperforms \cite{6araujo2017classification} by 8.3\% , where \cite{6araujo2017classification} developed a deep plain CNN for patch-wise prediction and combined patch-wise predictions with different voting schemas. We evaluated the proposed model on the BIC dataset with the exactly identical dataset setting in \cite{6araujo2017classification} and report 86.1\% accuracy on 4-class classification. Extended comparable analysis is shown in Table.~\ref{result}. The proposed residual network brings an improvement of 3.8\% for 4-class classification from the plain DCNN. The proposed deep fusion network significantly improves the classification accuracy by 4.51\% on BIC dataset compared to the best voting schema, which demonstrates the effectiveness of the fusion model.

All experiments were performed using PyTorch libaray on a NVIDIA 1080Ti GPU. The inference time for classifying a single high resolution histology image took roughly 80ms.

\section{Conclusion}
In this paper, we propose a deep spatial fusion network that models the complex construction of discriminative features over patches and learns to correct the patch-wise prediction bias on high resolution histology image. Also, we propose an adapted patch-wise residual network that effectively extract hierarchical visual features from cellular-level to overall tissue organization. Unlike previous patch-based CNN methods, the proposed architecture explores the spatial relationship between patches. Experiment results show that a substantially better performance than previous work even without using the ensemble. %We provide the PyTorch-based implementation on Github [6], regarding that the propose architecture can be applied to many other classification tasks on high resolution medical images. 

In future work, we plan to extend the current work by: (1) incorporating concise patch-wise feature maps on spatially organized probability maps, (2) employing the deep spatial fusion model to annotate malignant patches to assist diagnosis, and (3) transferring the proposed model to other high resolution medical images.

\bibliographystyle{splncs03}
\bibliography{ref}

\end{document}